\documentclass{article}
\usepackage{ijcai20}

\usepackage{times}
\usepackage{soul}
\usepackage{url}
\usepackage[utf8]{inputenc}
\usepackage[small]{caption}
\usepackage{graphicx}
\usepackage{amsmath}
\usepackage{amssymb}
\usepackage{amsthm}
\usepackage{booktabs}
\usepackage[english]{babel}
\usepackage{algorithm,algorithmicx,algpseudocode}
\usepackage{natbib}
\usepackage{svg}
\usepackage{subfig}
\usepackage{color}
\usepackage{etoolbox}
\usepackage{afterpage}
\usepackage{amsmath}
\usepackage{amssymb}
\usepackage{amsthm}
\usepackage{array}
\usepackage{booktabs}
\usepackage{caption}
\usepackage{enumitem}
\usepackage{fixme}
\usepackage{graphicx}
\usepackage{mathtools}
\usepackage{microtype}
\usepackage{multirow}
\usepackage{setspace}
\usepackage{subfig}
\usepackage{tabu}
\usepackage{tabularx}
\usepackage{tabulary}
\usepackage{titlesec}
\usepackage{xfrac}

\usepackage{amsmath,amsfonts,bm}

\def\eqref#1{equation~\ref{#1}}

\def\1{\bm{1}}

\DeclareMathAlphabet{\mathsfit}{\encodingdefault}{\sfdefault}{m}{sl}
\SetMathAlphabet{\mathsfit}{bold}{\encodingdefault}{\sfdefault}{bx}{n}

\DeclareMathOperator*{\argmax}{arg\,max}

\newcommand{\algoname}[1]{\textnormal{\textsc{#1}}}

\newcommand{\piekd}{PIEKD}

\title{Periodic Intra-Ensemble Knowledge Distillation for Reinforcement Learning}

\author{%
    Zhang-Wei Hong$^{1,2}$\footnote{This work was done during an internship at Preferred Networks.}\and
    Prabhat Nagarajan$^2$\and
    Guilherme Maeda$^2$
    \affiliations
    $^1$National Tsing Hua University\\
    $^2$Preferred Networks\\
    \emails
    \texttt{williamd4112@gapp.nthu.edu.tw},
   \texttt{\{prabhat,gjmaeda\}@preferred.jp}
}

\begin{document}

\maketitle

\begin{abstract}
Off-policy ensemble reinforcement learning (RL) methods have demonstrated impressive results across a range of RL benchmark tasks. Recent works suggest that directly imitating  experts' policies in a supervised manner before or during the course of training enables faster policy improvement for an RL agent. Motivated by these recent insights, we propose \textit{Periodic Intra-Ensemble Knowledge Distillation} (\piekd{}). \piekd{} is a learning framework that uses an ensemble of policies to act in the environment while periodically sharing knowledge amongst policies in the ensemble through knowledge distillation. Our experiments demonstrate that \piekd{} improves upon a state-of-the-art RL method in sample efficiency on several challenging MuJoCo benchmark tasks. Additionally, we perform ablation studies to better understand \piekd{}.
\end{abstract}

\section{Introduction}
\label{sec::intro}

In reinforcement learning (RL), the goal is to train a policy to interact with an environment, such that this policy yields the maximal expected return. While typical RL methods merely train a single parameterized policy, ensemble methods that share experiences amongst several function approximators~\citep{osband2017deep,osband2016deep} have been able to achieve superior performance in the context of reinforcement learning (RL). Unlike typical RL methods, \citet{osband2017deep} train an ensemble of neural network (NN) policies with distinct initial weights (i.e. parameters of NNs) simultaneously, by sharing experiences amongst the policies. These shared experiences are collected by first randomly selecting a policy from the ensemble to perform an episode. This episode of experiences is added to a shared experience replay buffer~\citep{mnih2015human} used to train all members of the ensemble. Learning from shared experience allows for more efficient policy learning, since randomly initialized policies result in extensive exploration in the environment. Though reinforcement learning from shared experiences has shown considerable improvement over single-policy RL methods, other lines of work~\citep{hester2018dqnfd} show that directly imitating an expert's experiences in a supervised manner can accelerate reinforcement learning.

\begin{figure}[htb!]
    \centering
    \includegraphics[width=0.48\textwidth]{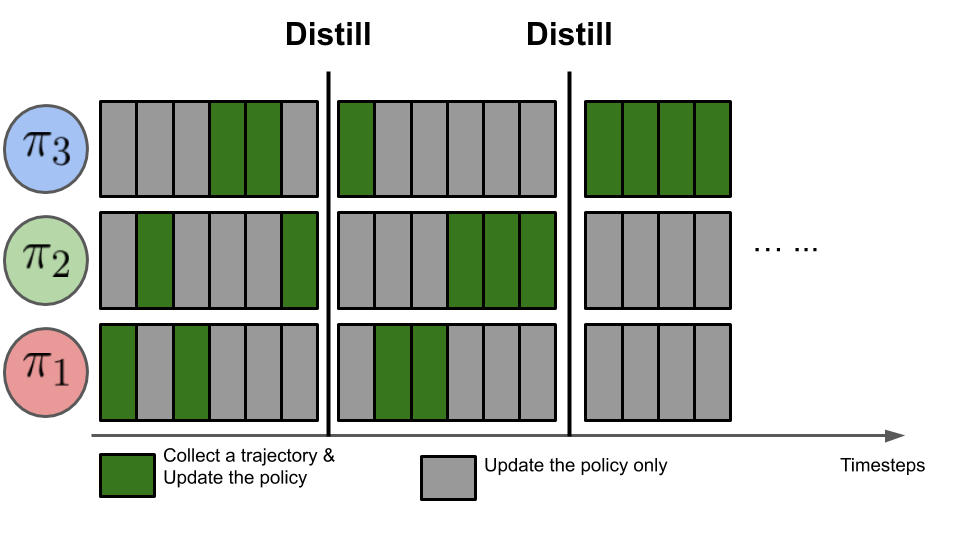}
    \caption{An overview of Periodic Intra-Ensemble Knowledge Distillation. We select a policy from the ensemble to act in the environment, and use this experience to update all policies. Periodically, we distill the best-performing policy to the rest of the ensemble.}
    \label{fig::workflow}
\end{figure}

Motivated by these results that demonstrate that direction imitation can accelerate RL, we propose Periodic Intra-Ensemble Knowledge Distillation (\piekd{}), a framework that not only trains an ensemble of policies via common experience but also shares the knowledge of the best-performing policy amongst the ensemble. Previous works on ensemble RL have shown that randomly initialized policies can result in adequate behavioral diversity~\citep{osband2016deep}. Thus \piekd{} first begins by initializing each policy in the ensemble with different weights to perform extensive exploration in the environment. As the behaviors of these policies are diverse in nature, at any given time during the course of training, one policy is naturally superior to other policies. This policy is then used to improve the quality of the other policies in the ensemble, without having to improve solely through experience. To use the best policy to improve other policies, \piekd{} employs knowledge distillation~\citep{hinton2015distilling}, which is effective at transferring knowledge between neural networks. By using knowledge distillation, we can encourage policies in the ensemble to act in a manner similar to the best policy, enabling them to rapidly improve and continue optimizing for the optimal policy from better starting points. Prior work~\citep{rusu2015policy} has shown that we can successfully distill several specialized policies into a single multitask policy, demonstrating that distillation can successfully augment behaviors into a policy without destroying existing knowledge. These results suggest that in \piekd{}, despite the use of distillation between policies', their inherent knowledge is still preserved, improving individual policies without destroying the diversity amongst policies. An abstract overview of \piekd{} is depicted in Figure~\ref{fig::workflow}.

This paper's primary contribution is Periodic Intra-Ensemble Knowledge Distillation (\piekd{}), a simple yet effective framework for off-policy RL that jointly trains an ensemble of policies while periodically performing knowledge sharing. We demonstrate empirically that \piekd{} can improve the state-of-the-art soft-actor critic (SAC)~\citep{haarnoja2018soft} on a suite of challenging MuJoCo tasks, exhibiting superior sample efficiency. We further validate the effectiveness of distillation for knowledge sharing by comparing against other forms of sharing knowledge.

The remainder of this paper is organized as follows. Section~\ref{sec::related_works} discusses the related work in ensemble RL and knowledge distillation. Section~\ref{sec::bg} provides a brief overview of the reinforcement learning formulation. Section~\ref{sec::method} describes \piekd{}. Section~\ref{sec::exp} presents our experimental findings. Lastly, Section~\ref{sec::conclusion} summarizes our contributions and outlines potential avenues for future work.
\section{Related work}
\label{sec::related_works}

The works that are most related to \piekd{}~\citep{osband2016deep,osband2017deep} train multiple policies via shared experience for the same task through RL, where the shared experiences are collected by all policies in the ensemble and stored in a common buffer, as our method does. Differing from those works~\citep{osband2016deep,osband2017deep}, we additionally periodically performing knowledge distillation between policies of the ensemble. Other related methods aggregate multiple policies to select actions~\citep{gimelfarb2018reinforcement,tham1995reinforcement}. \citet{abel2016exploratory} sequentially train a series of policies, boosting the learning performance by using the errors of a prior policy. However, rather than perform decision aggregation or sequentially-boosted training, we focus on improving the performance of each individual policy via knowledge sharing amongst jointly trained policies.

\citet{rusu2015policy} train a single neural network to perform multiple tasks by transferring multiple pretrained policies to a single network through distillation. \citet{hester2018dqnfd} and \citet{nair2018overcoming} accelerate RL agents' training progress through human experts' guidance. Rather than experts' policies, \citet{nagabandi2018mpc}, \citet{levine2013gps} and \citet{zhang2016mpc} leverage model-based controllers' behaviors, facilitating training for RL agents. Additionally, \citet{oh2018self} train RL agents to imitate past successful self-experiences or policies. Orthogonal to the aforementioned works, \piekd{} periodically exploits the current best policy within the ensemble, and shares its amongst the ensemble.

In other machine learning areas, \citet{zhang2018deep} trains multiple models that mutually imitate each other's outputs on classification tasks. Our distillation procedure is not mutual, but flows in a single direction, from a superior teacher policy to other student policies in the ensemble. Subsequent work by \citet{lan2018knowledge} trains an ensemble of models to imitate a stronger teacher model that aggregates all of the ensemble models' predictions. Our method contrasts from the above methods by periodically electing the teacher for distillation to other ensemble members. We maintain the distinction between ensemble members rather than aggregate them into a single policy. 

\citet{teh2017distral} and \citet{ghosh2017divide} distill multiple task-specific policies to a central multi-task policy and constrain the mutual divergence between each task-specific policy and the central one. \citet{galashov2019information} learn a task-specific policy while bounding the divergence between this task-specific policy and some generic policy that can perform basic task-agnostic behaviors. \citet{czarnecki2018mix} gradually transfer the knowledge of a simple policy to a complex policy during the course of joint training. Our work differs from the aforementioned works in several aspects. First, our method periodically elects a teacher policy for sharing knowledge rather than either constraining the mutual policy divergence~\citep{teh2017distral,ghosh2017divide,galashov2019information}. Second, our method does not rely on training heterogeneous policies (e.g. a simple policy and a complex policy~\citep{czarnecki2018mix}), which makes our method more generally applicable. Finally, as opposed to \citet{teh2017distral} and \citet{ghosh2017divide}, we consider single-task settings rather than multi-task settings.

Population-based methods similarly employ multiple policies in separate copies of environments to find the optimal policy. Evolutionary Algorithms (EA)~\citep{salimans2017evolution,gangwani2017policy,khadka2018evolution} randomly perturb the parameters of policies in the population, eliminate underperforming policies by evaluating the policies' performances in the environment, and produce new generations of policies from the remaining policies. Unlike EA, our method does not rely on separate copies of environments and eliminating existing policies from the population. Instead, our method focuses on continuously improving the existing policies. In addition to EA, other work~\citep{Jung2020Population-Guided} done concurrent to our work adds a regularization term that forces each agent to imitate the best agent's policy when performing policy updates at each step. Differing from \piekd{}, they train multiple agents in separate copies of the environment in parallel. Without the reliance on multiple copies of the environment, our method is more applicable in the cases of expensive interaction with the environment or costly setup of multiple environments (e.g. robot learning in the real world). 

\section{Background}
\label{sec::bg}

In this section we describe the general framework of RL. RL formalizes a sequential decision-making task as a \textit{Markov decision process} (MDP)~\citep{sutton1998introduction}. An MDP consists of a state space $\mathcal{S}$, a set of actions $\mathcal{A}$, a (potentially stochastic) transition function $\mathcal{T}: \mathcal{S} \times \mathcal{A} \rightarrow \mathcal{S}$, a reward function $\mathcal{R} : \mathcal{S} \times \mathcal{A} \rightarrow \mathbb{R}$, and a discount factor $\gamma \in [0,1]$. An RL agent performs episodes of a task where the agent starts in a random initial state $s_{0}$, sampled from the initial state distribution $\rho_{s_0}$, and performs actions, which transition the agent to new states and for which the agent receives rewards. More generally, at timestep $t$, an agent in state $s_{t}$ performs an action $a_{t}$, receives a reward $r_{t}$, and transitions to a new state $s_{t+1}$, according to the transition function $\mathcal{T}$. The discount factor $\gamma$ is used to indicate the agent's preference for short-term rewards over long-term rewards.

An RL agent performs actions according to its policy, a conditional probability distribution $\pi_{\phi}:\mathcal{S} \times \mathcal{A} \mapsto [0, 1]$, where $\phi$ denotes the parameters of the policy, which may be the parameters of a neural network. RL methods iteratively update $\phi$ via rollouts of experience $\tau = \{(s_t, a_t, r_t, s_{t+1})\}^{T-1}_{t=0}$, seeking within the parameter space $\Phi$ the optimal $\phi^*$ that maximizes the expected return $\mathbb{E}_{s \sim \rho_{s_0}}\big[ \sum^{T-1}_{t = 0} \gamma^{t} r_t | s_0 = s \big]$ at each $t$ within an episode.
\section{Method}
\label{sec::method}

In this section, we formally present the technical details of our method, Periodic Intra-Ensemble Knowledge Distillation (\piekd{}). We start by providing an overview of \piekd{} and then describe its components in detail.

\begin{algorithm}
  \caption{Periodic Intra-Ensemble Knowledge Distillation for Off-policy Actor Critic}
  \label{alg::ensemble_distill}
  \begin{algorithmic}[1]
  \Require an environment $\mathcal{E}$, an off-policy actor-crtic method $\mathcal{\omega}$, an ensemble size $K$, a parameter space $\Phi$, a set of parameterized policies and critics $\{\pi_{\phi_k}\}^{K-1}_{k=0}$ and $\{Q_{\theta_k}\}^{K-1}_{k=0}$, recent episodic performance statistics $\{R_k\}^{K-1}_{k=0}$, an episode length $T$, a distillation interval $I$, an experience buffer $\mathcal{D}$
  \State
  \State \textbf{i. Ensemble initialization}
  \State $\phi_k \sim \text{Uniform}(\Phi), \forall k \in [0, K)$
  \State $\mathcal{D} \leftarrow \{\}$
  \State  $R_{k} \leftarrow \{\}, \forall k \in [0, K)$
  \State  $t_{acc} \leftarrow 0$
  \While{not converged}
        \State \textbf{ii. Joint training}
        \State $k_{e} \sim \text{Uniform}([0, K))$  \Comment{Policy selection}
        \State $\tau \leftarrow \algoname{Rollout}(\mathcal{E}, \pi_{\phi_{k_{e}}})$
        \State $\mathcal{D} \leftarrow \mathcal{D} \cup \tau$
        \State \algoname{UpdatePolicy}($\pi_{\phi_k}$, $\mathcal{D}$, $\omega$), $\forall k \in [0, K)$
        \State \algoname{UpdateCritic}($Q_{\theta_k}$, $\mathcal{D}$, $\omega$), $\forall k \in [0, K)$
        \State \algoname{UpdateStat}$(R_{k_{e}}, \tau)$ \Comment{Update statistics}
        \State $t_{acc} \leftarrow t_{acc} + T$
        
        \State \textbf{iii. Intra-Ensemble Knowledge Distillation} 
        \If{$t_{acc} \ge I$}
            \State $k_{t} \leftarrow \mathop{\text{argmax}}_{k} {R_{{k}}}$ \Comment{Teacher election}
            \State \algoname{DistillPolicy}($\phi_k$, $\phi_{k_t}$, $\mathcal{D}$)$,\forall k \in [0, K)$ (Eq.~\ref{eq::policy_distill})
            \State \algoname{DistillCritic}($\theta_k$, $\theta_{k_t}$, $\mathcal{D}$)$,\forall k \in [0, K)$ (Eq.~\ref{eq::critic_distill})
            \State $t_{acc} \leftarrow 0$
        \EndIf
  \EndWhile
  \end{algorithmic}
\end{algorithm}

\subsection{Overview}
\label{subsec::method::overview}
\begin{figure*}[tb!]
    \includegraphics[width=\textwidth]{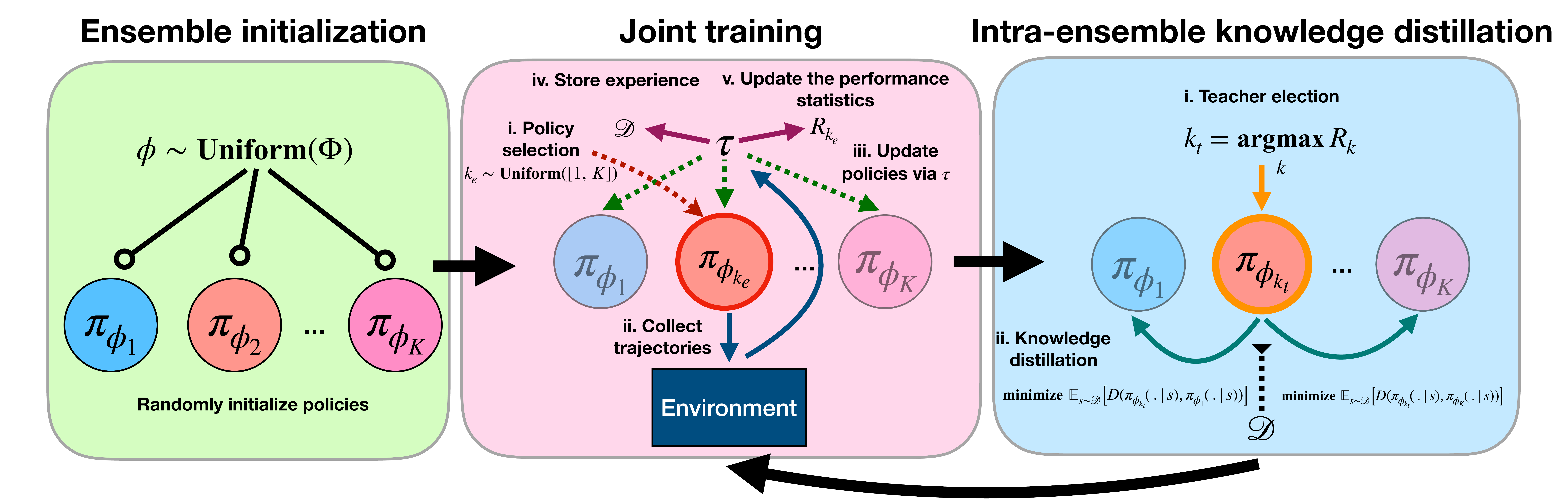}
    \caption{An overview of of the three phases of periodic intra-ensemble knowledge distillation: ensemble initialization, joint training, and intra-ensemble knowledge distillation.}
    \label{fig::overview}
\end{figure*}

\piekd{} maintains an ensemble of policies that perform that collect different experiences on the same task, and then periodically shares knowledge amongst the policies in the ensemble. \piekd{} is separated into three phases: ensemble initialization, joint training, and intra-ensemble knowledge distillation. First, the ensemble initialization phase randomly initializes an ensemble of policies with different parameters to achieve behavioral diversity. In the joint training stage, a policy randomly selected from the ensemble is used to execute an episode in the environment and its experience is then stored in a shared experience replay buffer that is used to train each policy. In the last stage, we perform intra-ensemble knowledge distillation, where we elect a teacher policy from the ensemble used to guide the other policies towards better behaviors. To this end, we distill~\citep{hinton2015distilling} the best-performing policy to the others. Algorithm \ref{alg::ensemble_distill} and Figure~\ref{fig::overview} summarize our method. In this paper, we apply \piekd{} to the state-of-the-art off-policy RL algorithm, soft actor-critic (SAC)~\citep{haarnoja2018learning}.

\subsection{Ensemble initialization}
\label{subsec::method::swarm}
In the ensemble initialization phase, we randomly initialize $K$ policies in the ensemble. Each policy is instantiated with a model parameterized by $\phi_k$, where $k$ stands for the policy's index in the ensemble. $\phi_k$ is initialized by sampling from the uniform distribution over parameter space $\Phi$ which contains all possible values of $\phi_k$: $\phi_k \sim \text{Uniform}(\Phi)$. Despite the simplicity of uniform distributions used for initialization, \citet{osband2016deep} shows that uniformly random initialization can provide adequate behavioral diversity. In this paper, we represent each $\phi_k, \forall k \in [0, K)$ as a neural network (NN), though other parametric models can be used.

Since SAC learns both a policy and a critic function that values states or state-action pairs from past experiences stored in a replay buffer~\citep{mnih2015human}, we create a shared replay buffer for all policies in the ensemble and randomly initialize a NN critic function $Q_{\theta_k}$ for each policy $\pi_{\phi_k}$. $\theta_k$ stands for the NN's weight for the critic $Q_{\theta_k}$.

\subsection{Joint training}
\label{subsec::method::exploration}
Each joint training phase consists of $I$ timesteps. For each episode, we select a policy in the ensemble to act in the environment (hereinafter, we refer this process as ``policy selection'') The policy selection strategy is a way of selecting a policy $\pi_{\phi_{k_{e}}}$ from the ensemble to perform an episode $\tau$ in the environment. This episode $\tau$ is stored in a shared experience replay buffer $\mathcal{D}$, and the policy's recent episodic performance statistic $R_{{{k_{e}}}}$ is updated according to the return achieved in $\tau$, where $R_{{{k_{e}}}}$ is the average episodic return in the most recent $M$ episodes. The episodic performance statistics $\{R_{{k}}\}^K_{k=0}$ and $\mathcal{D}$ will later be used in the intra-ensemble distillation phase. (Section~\ref{subsec::method::knowledgesharing}). In this paper, we adopt a simple uniform random policy selection strategy: $k_{e} \sim \text{Uniform}([0, K))$. To perform RL updates on the agent's policy,

After selecting a policy $\pi_{\phi_{k_{e}}}$ which performs an episode $\tau$, we store this $\tau$ in $\mathcal{D}$ (line 11). Then, we can sample data from $\mathcal{D}$  and update all policies and critics using SAC (line 12-13). Since off-policy RL methods like SAC do not require that $\tau$ is necessarily generated by the policy that is being updated, they enable our policies to learn from the trajectories generated by other policies of the ensemble. The details of the update routine for the policy and the critic are taken from the original SAC paper ~\citep{haarnoja2018soft}.

\subsection{Intra-ensemble knowledge distillation}
\label{subsec::method::knowledgesharing}
The intra-ensemble knowledge distillation phase consists of two stages: \textit{teacher election} and \textit{knowledge distillation}. The teacher election stage (line 18) selects a policy from the ensemble to serve as the teacher for other policies. In our experiments, we use the natural selection criteria of the selecting the best-performing teacher. Specifically, we select the policy that has the highest average recent episodic performance recorded in the joint training phase (Sec.~\ref{subsec::method::exploration}), namely $k_t = \argmax_k{R_k}$, where $k_t$ is the index of the teacher. Rather than use a policy's most recent episodic performance, we use its average return over its previous $M$ episodes, to minimize the noise in our estimate of the policy's performance.

Next, the elected teacher guides the other policies in the ensemble towards better policies (line 19-20). This is done through knowledge distillation~\citep{hinton2015distilling}, which has been shown to be effective at guiding a neural network to behave similarly to another. To distill from the teacher to the students (i.e., other policies in the ensemble), the teacher samples experiences from the buffer $\mathcal{D}$ and instructs each student to match the teacher's outputs on these samples. After distillation, the students acquire the teacher's knowledge, enabling them to correct their low-rewarding behaviors and reinforce their high-rewarding behaviors, without forgetting their previously learned behaviors~\citep{rusu2015policy,teh2017distral}. Specifically, the policy distillation process is formalized as updating each $\phi_k$ in the direction of
\begin{equation}
\label{eq::policy_distill}
    \nabla_{\phi_k} \mathbb{E}_{s \sim \mathcal{D}}\big[ D_{KL}(\pi_{\phi_{k_{t}}}(.|s) || \pi_{\phi_k}(.|s)) \big],
\end{equation}
where Kullback–Leibler divergence ($D_{KL}$) is a principled way to measure the similarity between two probability distributions (i.e., policies). Note that when applying \piekd{} to SAC, we must additionally distill the critic function from the teacher to the students, where each critic function is updated toward the direction
\begin{equation}
\label{eq::critic_distill}
    \nabla_{\theta_k} \mathbb{E}_{(s, a) \sim \mathcal{D}}\big[ (Q_{\theta_{k_{t}}}(s, a) - Q_{\theta_k}(s, a))^2 \big],
\end{equation}
where $\theta_k$ and $ \theta_{k_{t}}$ denote parameters of critic functions. $Q_{\theta_{k_{t}}}$ and $Q_{\theta_k}$ denote the critic function corresponding to the teacher's policy and the student's policy, respectively.

\section{Experiments}
\label{sec::exp}

The experiments are designed to answer the following questions: (1) Can \piekd{} improve upon the data efficiency of state-of-the-art RL? (2) Is knowledge distillation effective at sharing knowledge? (3) Is is it necessary to choose the best-performing agent to be the teacher? Next, we show our experimental findings for each of the aforementioned questions, and discuss their implications.

\subsection{Experimental setup}
\label{subsec::exp_setup}
\paragraph{Implementation.} ~ Our goal is to demonstrate how \piekd{} improves the sample efficiency of an RL algorithm. Since soft actor-critic (SAC)~\citep{haarnoja2018soft} exhibits state-of-the-art performance across several continuous control tasks, we build on top of SAC. We directly use the hyperparameters for SAC from the original paper~\citep{haarnoja2018soft} in all of our experiments\footnote{Code:\url{https://github.com/pfnet-research/piekd}}. Unless stated otherwise, the hyperparameters used in for \piekd{} (Algorithm~\ref{alg::ensemble_distill}) are $I = 5000$, and $K=3$. The value of $I$ is tuned via grid search over $[1000, 2000, \cdots, 10000]$. We tried different ensemble size configurations ($K \in \{2,3,5\}$) and found decided on $K=3$. For the remainder of our experiments, we term \piekd{} applied to SAC as \textit{SAC-\piekd{}}.

\paragraph{Benchmarks.} ~ We use OpenAI gym~\citep{openaigym}'s MuJoCo benchmark tasks, as used in the original SAC~\citep{haarnoja2018soft} paper. We choose most of the tasks selected in the original paper~\citep{haarnoja2018soft} to evaluate the performance of our method. The description for each task can be found in the source code for OpenAI gym \citep{openaigym}.

\paragraph{Evaluation.} ~ We adapt the evaluation approach from the original SAC paper~\citep{haarnoja2018soft}. We train each agent for 1 million timesteps, and run 20 evaluation episodes after every 10000 timesteps (i.e., number of interactions with the environment), where the performance is the mean of these 20 evaluation episodes. We repeat this entire process across 5 different runs, each with different random seeds. We plot the mean value and confidence interval of mean episodic return at each stage of training. The mean value and confidence interval are depicted by the solid line and shaded area, respectively. The confidence interval is estimated by the bootstrapped method. At each evaluation point, we report the highest mean episodic return amongst the agents in the ensemble. In some curves, we additionally report the lowest mean episodic return amongst the agents in the ensemble.

\subsection{Effectiveness of \piekd{}}
\label{subsec::exp_overall_perf}
In order to evaluate the effectiveness of intra-ensemble knowledge distillation, we compare \textit{SAC-\piekd{}}, against two baselines: \textit{Vanilla-SAC} and \textit{Ensemble-SAC}. \textit{Vanilla-SAC} denotes the original SAC; \textit{Ensemble-SAC} is the analogous variant of \citet{osband2016deep}'s method for ensemble Q-learning, except on SAC. At its core, Osband's method involves an ensemble of policies that act with the environment and generate experiences. These experiences are then used to train the entire ensemble using an off-policy RL algorithm, such as Q-learning or off-policy actor-critic methods. Thus, our \textit{Ensemble-SAC} baseline denotes the training of an ensemble of policies through SAC while sharing knowledge amongst in the ensemble in a shared replay buffer. Effectively, \textit{Ensemble-SAC} is \textit{SAC-\piekd{}} without the intra-ensemble knowledge distillation phase. For both \textit{Ensemble-SAC} and \textit{SAC-\piekd} we set the ensemble size $K$ to be 3.

\begin{figure*}[htb!]
    \centering
    \includegraphics[width=\textwidth]{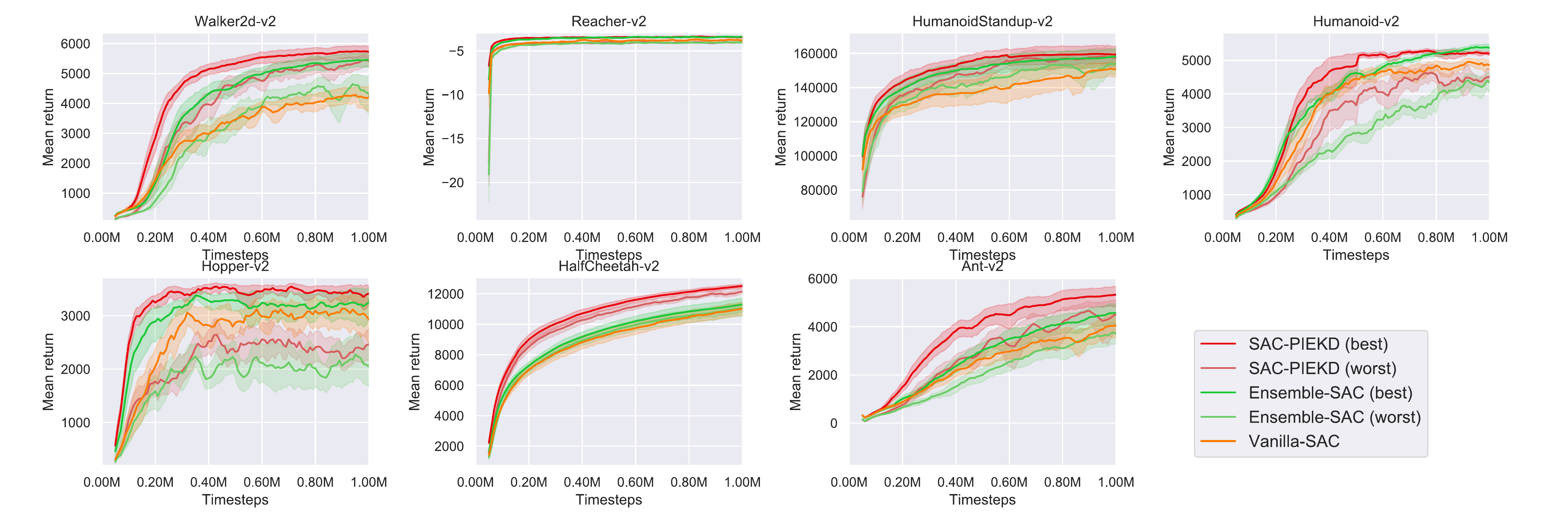}
    \caption{\textbf{Performance evaluation of \piekd{}}. \textit{SAC-\piekd{}} represents the implementation of our method upon SAC; \textit{Vanilla-SAC} stands for the original SAC; \textit{Ensemble-SAC} is an analogous variant of \citet{osband2016deep}'s method on \textit{Vanilla-SAC} (effectively \textit{SAC-\piekd{}} without intra-ensemble knowledge distillation). See Section~\ref{subsec::exp_overall_perf} for details. Notice that in most domains, \textit{SAC-\piekd{}} can reach the convergent performance of the baselines in less than half the training time.}
    \label{fig::overall_perf}
\end{figure*}

Our results are shown in Figure \ref{fig::overall_perf}. Note that we also plot the worst evaluation in the ensemble at each evaluation phase to provide insight into the general performance of the ensemble. In all tasks, we outperform all baselines, including \textit{Vanilla-SAC} and \textit{Ensemble-SAC}, in terms of sample efficiency. Visually we can see that throughout training, we have consistently better performance at similar amounts of experience, indicating that our method can achieve higher performance with the same number of experiences relative to our baselines.

\textit{SAC-\piekd{}} usually reaches the best baseline's convergent performance in half of the environment interactions. We even find that in the majority of tasks, our worst evaluation in the ensemble outperforms the baseline methods. This demonstrates that all policies of the ensemble are significantly improving, and our method's superior performance is not simply a consequence of selecting the best agent in the ensemble. In particular, \textit{SAC-\piekd{}}'s superiority over \textit{Ensemble-SAC} highlights the effectiveness of supplementing shared experiences (\textit{Ensemble-SAC}) with knowledge distillation. In summary, Figure \ref{fig::overall_perf} demonstrates the effectiveness of \piekd{} on enhancing the data efficiency of RL algorithms.

\subsection{Effectiveness of knowledge distillation for knowledge sharing}
\label{subsec::exp_extra}
In this section, we investigate the advantage of using knowledge distillation for knowledge sharing. We consider two alternative approaches towards sharing knowledge, other than distillation. First, we consider sharing knowledge by simply providing agents with additional policy updates (in lieu of distillation updates) using the shared experiences. We also consider directly copying the neural network as opposed to performing distillation. Below, we compare these two approaches against knowledge distillation.

\begin{figure*}[htb!]
    \centering
    \subfloat[\label{fig::extra}]{\includegraphics[width=0.33\linewidth]{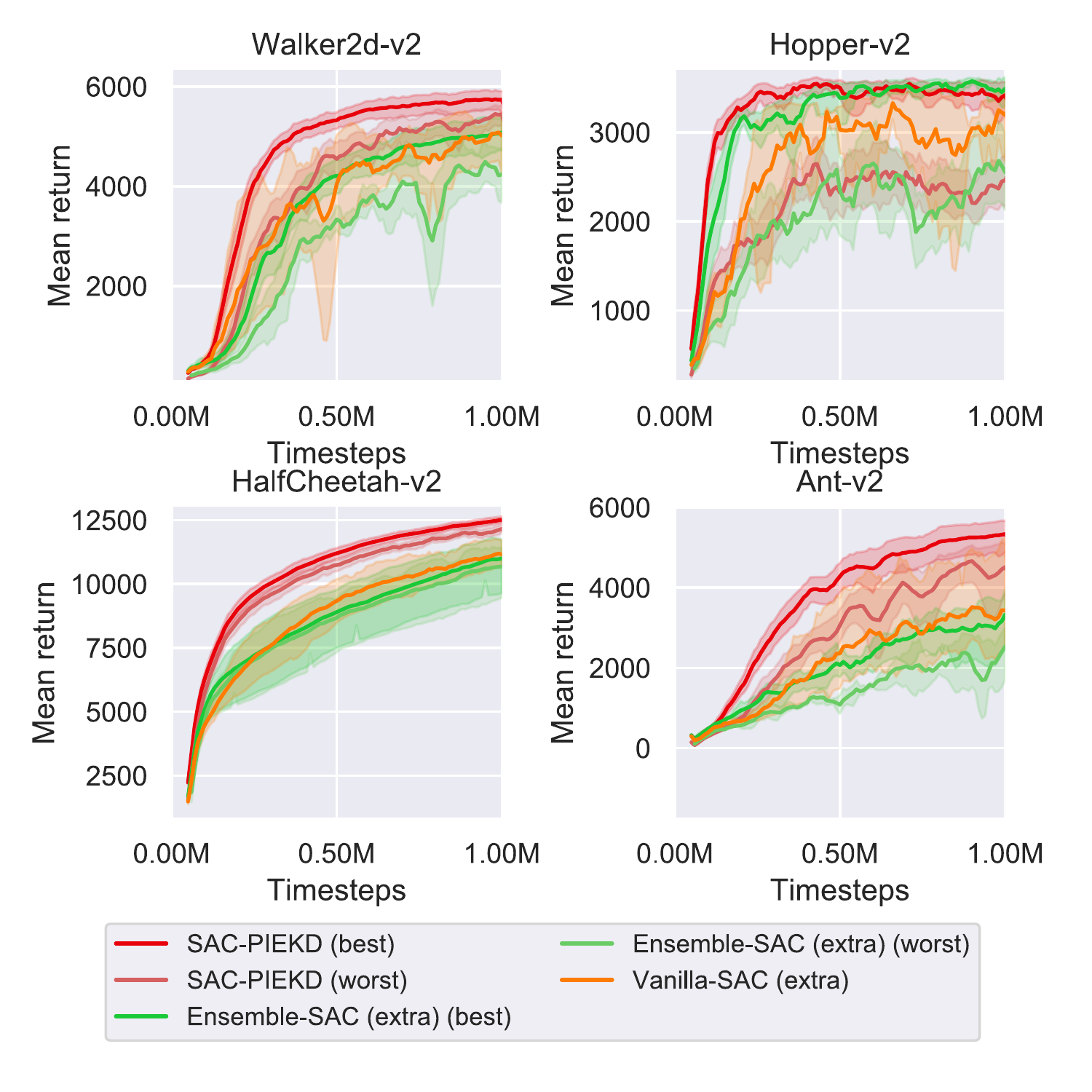}}
    \centering
    \subfloat[\label{fig::hardcopy}]{\includegraphics[width=0.33\linewidth]{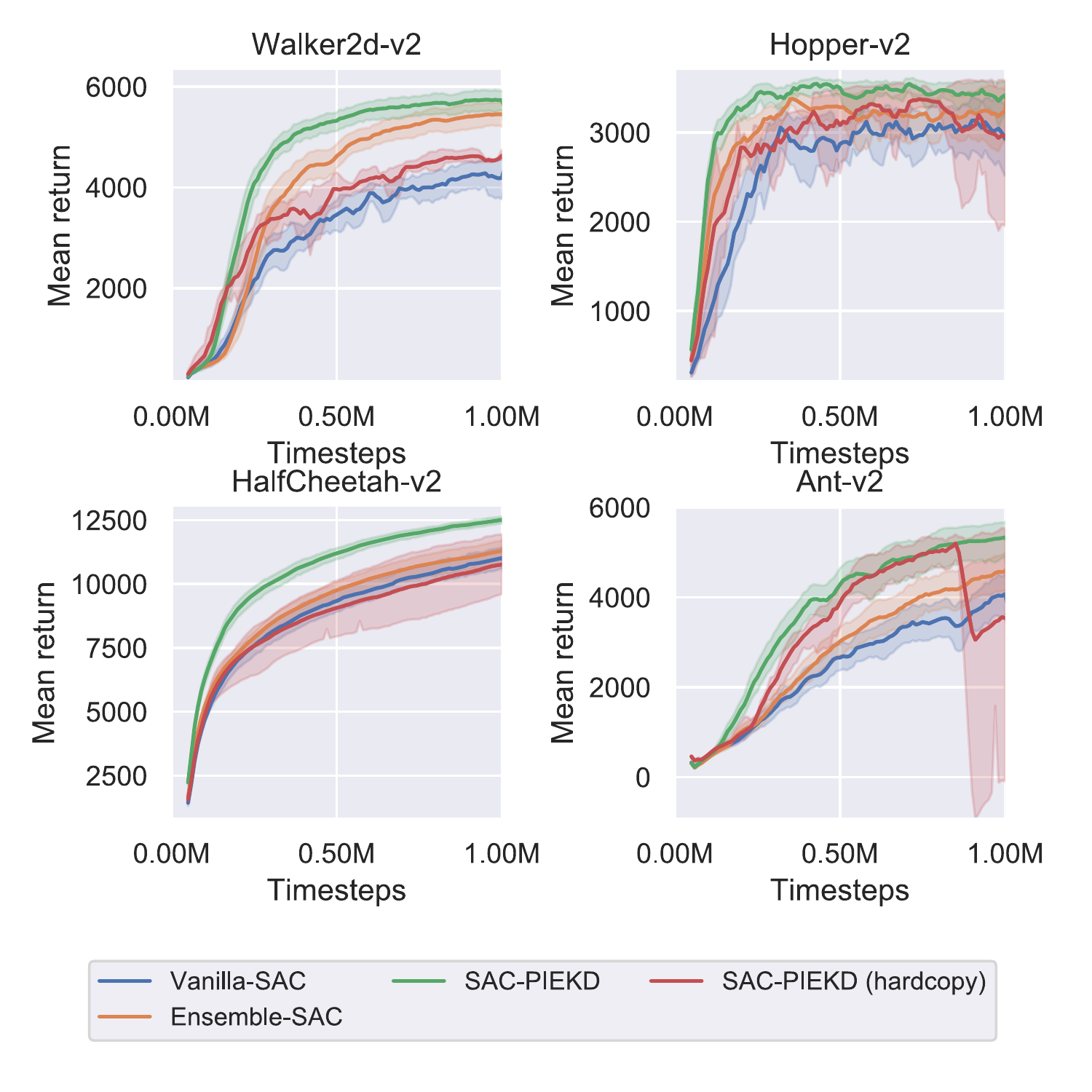}}
    \centering
    \subfloat[\label{fig::random_teacher}]{\includegraphics[width=0.33\linewidth]{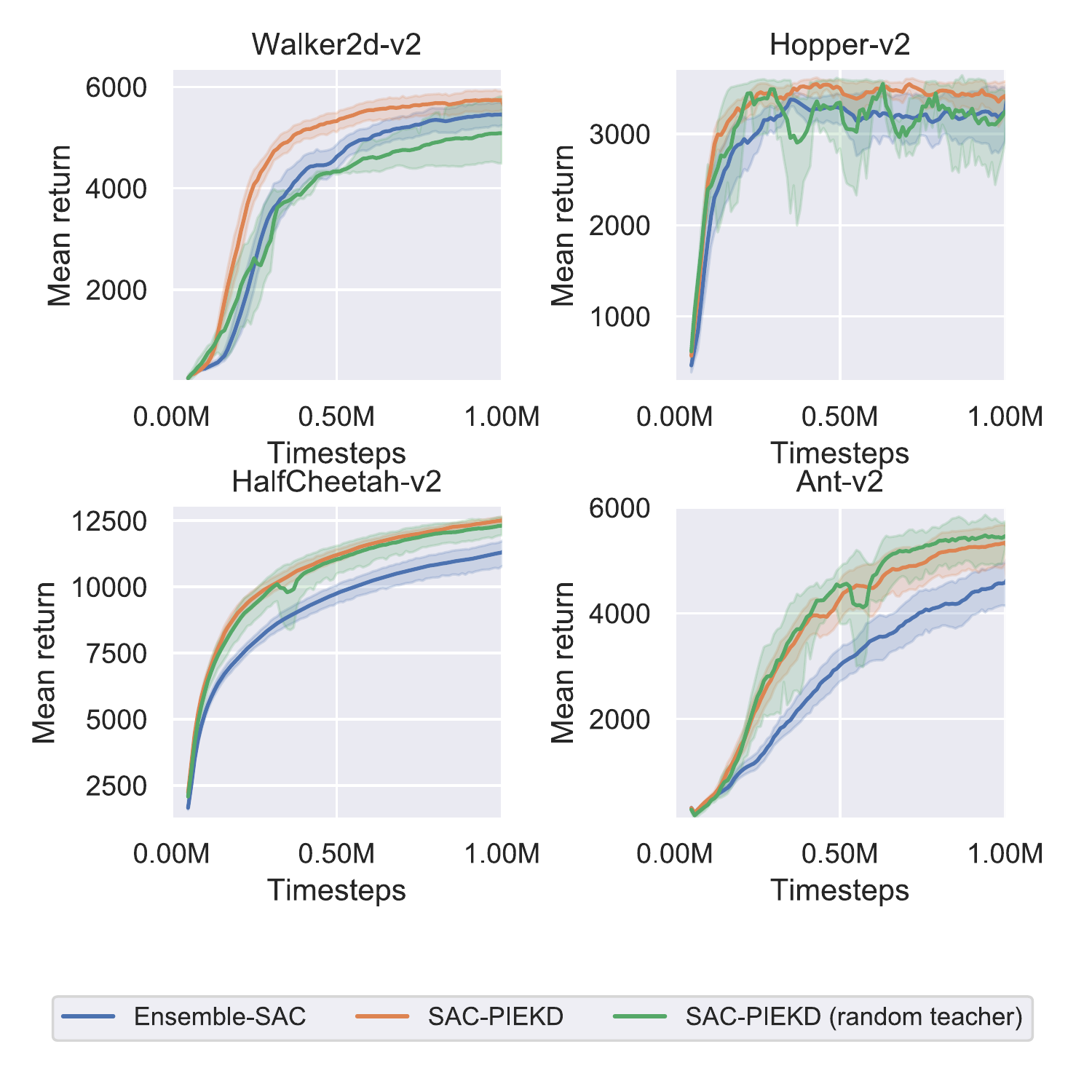}}
    \caption{(a) \textbf{Comparison between knowledge distillation and extra policy updates.} \textit{Vanilla-SAC (extra)} and \textit{Ensemble-SAC (extra)} stand for \textit{Vanilla-SAC} and \textit{Ensemble-SAC} variants that use extra policy updates, respectively (see Section~\ref{subsec::exp_extra} and Section~\ref{sec::bg} for details). (b) \textbf{Comparison between knowledge distillation and copying parameters.} \textit{SAC-\piekd{} (hardcopy)} stands for the variant of our method which directly copy the neural networks parameters of the best agent to the others. (c) \textbf{Comparison between the selecting the best-performing teacher vs. a random teacher.} \textit{SAC-\piekd{} (random teacher)} refers to the variant of our \textit{SAC-\piekd{}} where a randomly chosen teacher is used for knowledge distillation. This figure demonstrates that it can be more effective to select the best-performing agent as the teacher.}
\end{figure*}

Section~\ref{subsec::exp_overall_perf} has shown that \textit{Ensemble-SAC}, which updates all agents' policies through shared experiences fails to learn as efficiently as \textit{SAC-\piekd{}}. However, \textit{SAC-\piekd{}} uses additional gradient updates during knowledge distillation phase, whereas \textit{Ensemble-SAC} only performs joint training, and lacks an additional knowledge distillation phase. It is unclear whether additional policy updates in lieu of knowledge distillation can achieve the same effects. To investigate this, we compare \textit{SAC-\piekd{}} with \textit{Vanilla-SAC (extra)} and \textit{Ensemble-SAC (extra)}, which respectively correspond to \textit{Vanilla-SAC} and \textit{Ensemble-SAC} (see Section~\ref{subsec::exp_overall_perf}) that are trained with extra policy update steps with the same number of updates and minibatch sizes that \textit{SAC-\piekd{}} performs. A policy update here refers to a training step that updates the policy and value function~\citep{haarnoja2018soft}, if required, by RL algorithms. Figure~(\ref{fig::extra}) compares the performance of our baselines to \textit{SAC-\piekd{}}. We see that \textit{SAC-\piekd{}} reaches higher performance more rapidly than the baselines. This observation shows that knowledge distillation is more effective than policy updates for knowledge sharing.

We additionally study whether the naive method of directly copying parameters from the best-performing agent can also be an effective way to share knowledge between neural networks. We compare a variant of our method, which we denote as \textit{SAC-\piekd{} (hardcopy)}, against \textit{SAC-\piekd{}}. In \textit{SAC-\piekd{} (hardcopy)}, rather than perform intra-ensemble knowledge distillation, we simply copy the parameters of the teacher policy and critic into the student policies and critics. Figure~(\ref{fig::hardcopy}) depicts the performance of this variant. We see that \textit{SAC-\piekd{} (hardcopy)} performs worse than both \textit{Ensemble-SAC} and \textit{SAC-\piekd{}}. Thus, it is clear that knowledge distillation is superior to naively copying the best agent's parameters. In fact, it can be counterproductive to explicitly copy parameters, as \textit{Ensemble-SAC} outperforms copying without any knowledge sharing. This is likely due to the loss in policy diversity as a consequence of hardcopying, perhaps reducing to training a single policy as in \textit{Vanilla-SAC}.

\subsection{Effectiveness of selecting the best-performing agent as the teacher}
\label{subsec::random_teacher}
During teacher election, we opted for the natural strategy of selecting the best-performing agent. However, in order to investigate its importance, we compared the performance of \textit{SAC-\piekd{}} when we select the best policy to be the teacher as opposed to selecting a random policy to be the teacher. This is depicted in Figure~\ref{fig::random_teacher}, where \textit{SAC-\piekd{} (random teacher)} denotes the selection of a random policy to be the teacher and the standard \textit{SAC-\piekd{}} refers to the selection of the highest-performing policy to be the teacher. We see that using the highest-performing teacher for distillation appears to be slightly better than selecting a random teacher, though not significantly. Interestingly, we see that using a random teacher performs better than \textit{Ensemble-SAC}. This result suggests that selecting the best teacher is not necessarily of high importance, as a random teacher yields benefits. While this warrants further investigation, perhaps the diverse knowledge is being shared through distillation, which may elicit the success we see in \textit{SAC-\piekd{} (random teacher)}. Another possibility is that by bringing policies closer together, the off-policy error~\citep{fujimoto2018off} stemming from RL updates on a shared replay buffer is reduced, improving performance. However, we can conclude that selecting the highest-performing teacher, while somewhat beneficial, is nonessential, and we leave the investigation of these open questions for future work.
\section{Conclusion}
\label{sec::conclusion}
\vspace{-1ex}

In this paper, we introduce Periodic Intra-Ensemble Knowledge Distillation (\piekd{}), a method that jointly trains an ensemble of RL policies while periodically sharing information via knowledge distillation. Our experimental results demonstrate that \piekd{} improves the data efficiency of a state-of-the-art RL method on several standard MuJoCo tasks. Also, we show that knowledge distillation is more effective than the other approaches for knowledge sharing. We found that electing the best-performing policy is beneficial, but nonessential for improving the sample efficiency  of \piekd{}.

\piekd{} opens several avenues for future work. While we used a simple uniform policy selection strategy, a more efficient policy selection strategy may further accelerate learning. Moreover, while our ensemble members used identical architectures, \piekd{} may benefit from using heterogeneous ensembles, consisting of different architectures that may be conducive to learning different skills, which can then be distilled within the ensemble.  Lastly, additional investigations into teacher elections may be lead to informative insights.

\bibliographystyle{named}
\bibliography{ijcai20}

\begin{thebibliography}{}

\bibitem[\protect\citeauthoryear{Abel \bgroup \em et al.\egroup
  }{2016}]{abel2016exploratory}
David Abel, Alekh Agarwal, Fernando Diaz, Akshay Krishnamurthy, and Robert~E
  Schapire.
\newblock Exploratory gradient boosting for reinforcement learning in complex
  domains.
\newblock {\em arXiv preprint arXiv:1603.04119}, 2016.

\bibitem[\protect\citeauthoryear{Brockman \bgroup \em et al.\egroup
  }{2016}]{openaigym}
Greg Brockman, Vicki Cheung, Ludwig Pettersson, Jonas Schneider, John Schulman,
  Jie Tang, and Wojciech Zaremba.
\newblock Openai gym, 2016.

\bibitem[\protect\citeauthoryear{Czarnecki \bgroup \em et al.\egroup
  }{2018}]{czarnecki2018mix}
Wojciech~Marian Czarnecki, Siddhant~M Jayakumar, Max Jaderberg, Leonard
  Hasenclever, Yee~Whye Teh, Simon Osindero, Nicolas Heess, and Razvan Pascanu.
\newblock Mix\&match-agent curricula for reinforcement learning.
\newblock {\em arXiv preprint arXiv:1806.01780}, 2018.

\bibitem[\protect\citeauthoryear{Fujimoto \bgroup \em et al.\egroup
  }{2018}]{fujimoto2018off}
Scott Fujimoto, David Meger, and Doina Precup.
\newblock Off-policy deep reinforcement learning without exploration.
\newblock {\em arXiv preprint arXiv:1812.02900}, 2018.

\bibitem[\protect\citeauthoryear{Galashov \bgroup \em et al.\egroup
  }{2019}]{galashov2019information}
Alexandre Galashov, Siddhant~M Jayakumar, Leonard Hasenclever, Dhruva Tirumala,
  Jonathan Schwarz, Guillaume Desjardins, Wojciech~M Czarnecki, Yee~Whye Teh,
  Razvan Pascanu, and Nicolas Heess.
\newblock Information asymmetry in kl-regularized rl.
\newblock {\em arXiv preprint arXiv:1905.01240}, 2019.

\bibitem[\protect\citeauthoryear{Gangwani and Peng}{2017}]{gangwani2017policy}
Tanmay Gangwani and Jian Peng.
\newblock Policy optimization by genetic distillation.
\newblock {\em arXiv preprint arXiv:1711.01012}, 2017.

\bibitem[\protect\citeauthoryear{Ghosh \bgroup \em et al.\egroup
  }{2017}]{ghosh2017divide}
Dibya Ghosh, Avi Singh, Aravind Rajeswaran, Vikash Kumar, and Sergey Levine.
\newblock Divide-and-conquer reinforcement learning.
\newblock {\em arXiv preprint arXiv:1711.09874}, 2017.

\bibitem[\protect\citeauthoryear{Gimelfarb \bgroup \em et al.\egroup
  }{2018}]{gimelfarb2018reinforcement}
Michael Gimelfarb, Scott Sanner, and Chi-Guhn Lee.
\newblock Reinforcement learning with multiple experts: A bayesian model
  combination approach.
\newblock In {\em Advances in Neural Information Processing Systems}, pages
  9528--9538, 2018.

\bibitem[\protect\citeauthoryear{Haarnoja \bgroup \em et al.\egroup
  }{2018a}]{haarnoja2018learning}
Tuomas Haarnoja, Aurick Zhou, Sehoon Ha, Jie Tan, George Tucker, and Sergey
  Levine.
\newblock Learning to walk via deep reinforcement learning.
\newblock {\em arXiv preprint arXiv:1812.11103}, 2018.

\bibitem[\protect\citeauthoryear{Haarnoja \bgroup \em et al.\egroup
  }{2018b}]{haarnoja2018soft}
Tuomas Haarnoja, Aurick Zhou, Kristian Hartikainen, George Tucker, Sehoon Ha,
  Jie Tan, Vikash Kumar, Henry Zhu, Abhishek Gupta, Pieter Abbeel, et~al.
\newblock Soft actor-critic algorithms and applications.
\newblock {\em arXiv preprint arXiv:1812.05905}, 2018.

\bibitem[\protect\citeauthoryear{Hester \bgroup \em et al.\egroup
  }{2018}]{hester2018dqnfd}
Todd Hester, Matej Vecerik, Olivier Pietquin, Marc Lanctot, Tom Schaul, Bilal
  Piot, Dan Horgan, John Quan, Andrew Sendonaris, Ian Osband, et~al.
\newblock Deep q-learning from demonstrations.
\newblock In {\em Thirty-Second AAAI Conference on Artificial Intelligence},
  2018.

\bibitem[\protect\citeauthoryear{Hinton \bgroup \em et al.\egroup
  }{2015}]{hinton2015distilling}
Geoffrey Hinton, Oriol Vinyals, and Jeff Dean.
\newblock Distilling the knowledge in a neural network.
\newblock {\em arXiv preprint arXiv:1503.02531}, 2015.

\bibitem[\protect\citeauthoryear{Jung \bgroup \em et al.\egroup
  }{2020}]{Jung2020Population-Guided}
Whiyoung Jung, Giseung Park, and Youngchul Sung.
\newblock Population-guided parallel policy search for reinforcement learning.
\newblock In {\em International Conference on Learning Representations}, 2020.

\bibitem[\protect\citeauthoryear{Khadka and Tumer}{2018}]{khadka2018evolution}
Shauharda Khadka and Kagan Tumer.
\newblock Evolution-guided policy gradient in reinforcement learning.
\newblock In {\em Advances in Neural Information Processing Systems}, pages
  1188--1200, 2018.

\bibitem[\protect\citeauthoryear{Lan \bgroup \em et al.\egroup
  }{2018}]{lan2018knowledge}
Xu~Lan, Xiatian Zhu, and Shaogang Gong.
\newblock Knowledge distillation by on-the-fly native ensemble.
\newblock In {\em Proceedings of the 32nd International Conference on Neural
  Information Processing Systems}, pages 7528--7538. Curran Associates Inc.,
  2018.

\bibitem[\protect\citeauthoryear{Levine and Koltun}{2013}]{levine2013gps}
Sergey Levine and Vladlen Koltun.
\newblock Guided policy search.
\newblock In {\em International Conference on Machine Learning}, pages 1--9,
  2013.

\bibitem[\protect\citeauthoryear{Mnih \bgroup \em et al.\egroup
  }{2015}]{mnih2015human}
Volodymyr Mnih, Koray Kavukcuoglu, David Silver, Andrei~A Rusu, Joel Veness,
  Marc~G Bellemare, Alex Graves, Martin Riedmiller, Andreas~K Fidjeland, Georg
  Ostrovski, et~al.
\newblock Human-level control through deep reinforcement learning.
\newblock {\em Nature}, 518(7540):529, 2015.

\bibitem[\protect\citeauthoryear{Nagabandi \bgroup \em et al.\egroup
  }{2018}]{nagabandi2018mpc}
Anusha Nagabandi, Gregory Kahn, Ronald~S Fearing, and Sergey Levine.
\newblock Neural network dynamics for model-based deep reinforcement learning
  with model-free fine-tuning.
\newblock In {\em 2018 IEEE International Conference on Robotics and Automation
  (ICRA)}, pages 7559--7566. IEEE, 2018.

\bibitem[\protect\citeauthoryear{Nair \bgroup \em et al.\egroup
  }{2018}]{nair2018overcoming}
Ashvin Nair, Bob McGrew, Marcin Andrychowicz, Wojciech Zaremba, and Pieter
  Abbeel.
\newblock Overcoming exploration in reinforcement learning with demonstrations.
\newblock In {\em 2018 IEEE International Conference on Robotics and Automation
  (ICRA)}, pages 6292--6299. IEEE, 2018.

\bibitem[\protect\citeauthoryear{Oh \bgroup \em et al.\egroup
  }{2018}]{oh2018self}
Junhyuk Oh, Yijie Guo, Satinder Singh, and Honglak Lee.
\newblock Self-imitation learning.
\newblock {\em arXiv preprint arXiv:1806.05635}, 2018.

\bibitem[\protect\citeauthoryear{Osband \bgroup \em et al.\egroup
  }{2016}]{osband2016deep}
Ian Osband, Charles Blundell, Alexander Pritzel, and Benjamin Van~Roy.
\newblock Deep exploration via bootstrapped dqn.
\newblock In {\em Advances in neural information processing systems}, pages
  4026--4034, 2016.

\bibitem[\protect\citeauthoryear{Osband \bgroup \em et al.\egroup
  }{2017}]{osband2017deep}
Ian Osband, Benjamin Van~Roy, Daniel Russo, and Zheng Wen.
\newblock Deep exploration via randomized value functions.
\newblock {\em arXiv preprint arXiv:1703.07608}, 2017.

\bibitem[\protect\citeauthoryear{Rusu \bgroup \em et al.\egroup
  }{2015}]{rusu2015policy}
Andrei~A Rusu, Sergio~Gomez Colmenarejo, Caglar Gulcehre, Guillaume Desjardins,
  James Kirkpatrick, Razvan Pascanu, Volodymyr Mnih, Koray Kavukcuoglu, and
  Raia Hadsell.
\newblock Policy distillation.
\newblock {\em arXiv preprint arXiv:1511.06295}, 2015.

\bibitem[\protect\citeauthoryear{Salimans \bgroup \em et al.\egroup
  }{2017}]{salimans2017evolution}
Tim Salimans, Jonathan Ho, Xi~Chen, Szymon Sidor, and Ilya Sutskever.
\newblock Evolution strategies as a scalable alternative to reinforcement
  learning.
\newblock {\em arXiv preprint arXiv:1703.03864}, 2017.

\bibitem[\protect\citeauthoryear{Sutton \bgroup \em et al.\egroup
  }{1998}]{sutton1998introduction}
Richard~S Sutton, Andrew~G Barto, et~al.
\newblock {\em Introduction to reinforcement learning}, volume 135.
\newblock MIT press Cambridge, 1998.

\bibitem[\protect\citeauthoryear{Teh \bgroup \em et al.\egroup
  }{2017}]{teh2017distral}
Yee Teh, Victor Bapst, Wojciech~M Czarnecki, John Quan, James Kirkpatrick, Raia
  Hadsell, Nicolas Heess, and Razvan Pascanu.
\newblock Distral: Robust multitask reinforcement learning.
\newblock In {\em Advances in Neural Information Processing Systems}, pages
  4496--4506, 2017.

\bibitem[\protect\citeauthoryear{Tham}{1995}]{tham1995reinforcement}
Chen~K Tham.
\newblock Reinforcement learning of multiple tasks using a hierarchical cmac
  architecture.
\newblock {\em Robotics and Autonomous Systems}, 15(4):247--274, 1995.

\bibitem[\protect\citeauthoryear{Zhang \bgroup \em et al.\egroup
  }{2016}]{zhang2016mpc}
Tianhao Zhang, Gregory Kahn, Sergey Levine, and Pieter Abbeel.
\newblock Learning deep control policies for autonomous aerial vehicles with
  mpc-guided policy search.
\newblock In {\em 2016 IEEE international conference on robotics and automation
  (ICRA)}, pages 528--535. IEEE, 2016.

\bibitem[\protect\citeauthoryear{Zhang \bgroup \em et al.\egroup
  }{2018}]{zhang2018deep}
Ying Zhang, Tao Xiang, Timothy~M Hospedales, and Huchuan Lu.
\newblock Deep mutual learning.
\newblock In {\em Proceedings of the IEEE Conference on Computer Vision and
  Pattern Recognition}, pages 4320--4328, 2018.

\end{thebibliography}

\end{document}